\documentclass{article}

\usepackage[preprint]{neurips_2024}

\usepackage[utf8]{inputenc} % allow utf-8 input
\usepackage[T1]{fontenc}    % use 8-bit T1 fonts
\usepackage{url}            % simple URL typesetting
\usepackage{booktabs}       % professional-quality tables
\usepackage{multirow}
\usepackage{adjustbox}
\usepackage{amsfonts}       % blackboard math symbols
\usepackage{nicefrac}       % compact symbols for 1/2, etc.
\usepackage{microtype}      % microtypography
\usepackage[table]{xcolor}         % colors
\usepackage{listings}
\usepackage{enumitem}
\usepackage{parskip}
\usepackage{wrapfig}
\usepackage{pifont}
\usepackage{xspace}
\usepackage{amssymb}
\usepackage{subcaption}
\usepackage{ulem}
\usepackage{tcolorbox}
\usepackage{listings}
\usepackage{algorithm}  % For the algorithm floating environment
\usepackage{algpseudocode}  % For the algorithmic environment
\usepackage{algorithmicx}  % For algorithmic commands
\usepackage{amsmath}  % For mathematical symbols
\usepackage{textcomp}  % For special text symbols
\usepackage{caption}
\definecolor{softblue}{RGB}{100, 149, 237}
\definecolor{darkpurple}{RGB}{160, 0, 160}
\definecolor{darkred}{RGB}{160, 0, 0}
\definecolor{darkblue}{RGB}{0, 0, 160}
\usepackage[colorlinks,linkcolor=black,anchorcolor=blue,citecolor=darkblue]{hyperref}

\hypersetup{
colorlinks = true,
linkcolor = darkblue,
urlcolor = darkred,
citecolor = darkpurple,
filecolor = blue,
pdfauthor = {Author},
pdftitle = {Title},
pdfsubject = {subject},
pdfkeywords = {one, two},
pdfproducer = {LaTeX},
pdfcreator = {pdfLaTeX},
}

\definecolor{darkgreen}{rgb}{0.0, 0.5, 0.0}
\definecolor{darkred}{rgb}{0.5, 0.0, 0.0}

\newcommand{\platform}{\mbox{AIOS 1.0}\xspace}
\newcommand{\agent}{\mbox{LiteCUA}\xspace}

\title{\agent: Computer as MCP Server for Computer-Use Agent on AIOS}
\author{
  Kai Mei
  \And
  Xi Zhu
  \And
  Hang Gao
  \And
  Shuhang Lin
  \And
  Yongfeng Zhang
  \thanks{\textbf{Corresponding author}: Yongfeng Zhang, yongfeng.zhang@rutgers.edu, Department of Computer Science, Rutgers University, New Brunswick, NJ 08854}
}

\begin{document}

\maketitle

\begin{abstract}
We present \platform, a novel platform designed to advance computer-use agent (CUA) capabilities through environmental contextualization. While existing approaches primarily focus on building more powerful agent frameworks or enhancing agent models, we identify a fundamental limitation: the semantic disconnect between how language models understand the world and how computer interfaces are structured. \platform addresses this challenge by transforming computers into contextual environments that language models can natively comprehend, implementing a Model Context Protocol (MCP) server architecture to abstract computer states and actions. This approach effectively decouples interface complexity from decision complexity, enabling agents to reason more effectively about computing environments. To demonstrate our platform's effectiveness, we introduce \agent, a lightweight computer-use agent built on \platform that achieves a 14.66\% success rate on the OSWorld benchmark, outperforming several specialized agent frameworks despite its simple architecture. Our results suggest that contextualizing computer environments for language models represents a promising direction for developing more capable computer-use agents and advancing toward AGI that can interact with digital systems.

\end{abstract}

\section{Introduction} \label{sec:intro}

Large language models (LLMs) \citep{OpenAIGPT4, team2023gemini, liu2024deepseek, dubey2024llama, yang2024qwen2} have achieved impressive leaps in understanding knowledge and reasoning \citep{wei2023chainofthought, yao2023tree, besta2023graph, jin2024impact, jin2025two, chen2022program, jin2024learning}, fuelling a new generation of LLM-powered agents that plan, observe, and act to interact with digital environments \citep{shen2023hugginggpt, wu2023autogen, ge2023openagi, yao2023react}. Research demonstrates that, when coupled with suitable prompting or fine-tuning, LLM-based agents can effectively plan to fulfill complex tool calling \citep{patil2024gorilla, qin2023toolllm, xu2023tool} and programming \citep{wang2024executable, yangswe, zhang2024codeagent} tasks.

Moving beyond calling external tools or executing generated code, researchers have begun exploring whether LLM agents can perform more complex computer-use tasks, leveraging their emerging planning and action capabilities. This exploration has catalyzed the development of benchmarks assessing LLM-based agents in computer-use scenarios across operating systems \citep{xie2024osworld, bonatti2024windows}, mobile devices \citep{deng2024mobile, rawles2024androidworld}, and browsers \citep{yao2022webshop, zhou2023webarena, deng2023mind2web, koh2024visualwebarena, pan2024webcanvas}. Parallel efforts have focused on leveraging LLMs and vision-language models (VLMs) to perceive computer/mobile environments and execute actions such as clicking, typing, and dragging interactive elements to complete tasks. Current approaches to computer-use agents (CUAs) primarily follow two trajectories: (1) developing agent frameworks with specialized modules for critical subtasks like grounding \citep{agashe2025agent, gou2025navigating}, search \citep{yu2024exact}, and reflection \citep{zhang2025enhancing, wang2024oscar}; and (2) embedding agentic capabilities directly into LLMs/VLMs to create native agent models \citep{qin2025ui, guo2025seed1}. Despite these advances, CUAs continue to exhibit significant limitations in their ability to navigate complex computing environments effectively.

\begin{figure}
    \centering
    \includegraphics[width=1.0\linewidth]{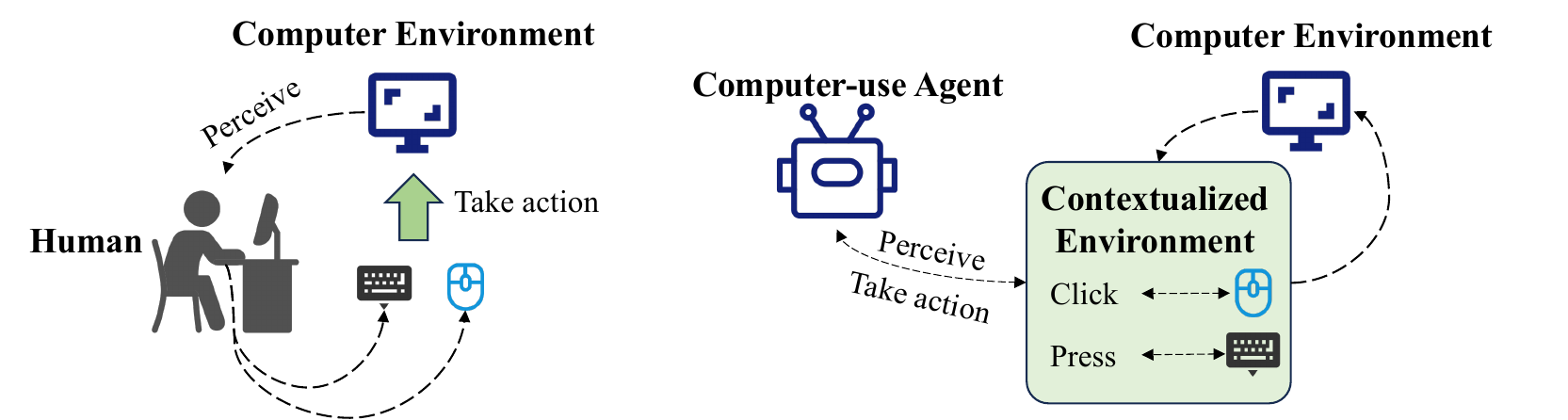}
    \caption{Illustration of the differences between how human operates computer and how computer-use agent (CUA) operates computer, where CUA requires a specific contextualized environment to understand the original computer environment and take actions. } \label{fig:intro}
    \label{fig:enter-label}
\end{figure}

These limitations stem from a fundamental challenge illustrated in \autoref{fig:intro}: a profound disconnect between how humans and AI agents perceive and interact with computing environments. In the current paradigm, researches on CUA predominantly focus on enhancing agent capabilities while largely overlooking the environments in which these agents operate. As the right part of \autoref{fig:intro} illustrates, there exists a significant opportunity to bridge this gap by creating agent-specific environments that mediate between the raw computer interface and the agent's cognition. Rather than forcing agents to adapt to interfaces designed for human perception and action, we can transform these environments into contextualized representations that align with the semantic understanding of language models.

Inspired by this and based on the foundation laid out by the AIOS architecture \citep{mei2024aios,rama2025cerebrum},  we introduce \textbf{\platform}, an augmented platform developed to move towards contextualization of computer environments for LLM-based agents. Specifically, \platform aims to model the full computer as MCP\footnote{https://modelcontextprotocol.io/introduction} server to provide LLM-friendly and well-structured context where interactive elements permissible actions are explicitly encoded in JSON schemas. This contextualization process transforms the computer from an opaque operational environment into a semantic landscape that aligns with the LLM's understanding of the environment. This design equips agents with an informative yet compact view of the state space, enabling finer-grained reasoning over available operations and facilitating longer-horizon planning. By decoupling interface complexity from decision complexity, \platform addresses a fundamental bottleneck in current CUA capabilities. The contextualized environment acts as a semantic bridge between the agent's language-based reasoning and the computer's operational requirements. 

Upon \platform, we introduce \textbf{\agent}, a computer-use agent that is developed by us with simple architectural design. \agent achieves competitive results with several agent frameworks, which scores 14.66\% in the OSWorld benchmark \citep{xie2024osworld}. The success of \agent emphasizes that \platform's approach of creating agent-specific environments can effectively reduce the implementation complexity required to build computer-use agents, allowing researchers to focus on advancing agent capabilities rather than wrestling with the intricacies of computer interface interpretation. By reframing the computer-use agent challenge as an environmental contextualization problem rather than solely a model capability issue, \platform aims to catalyze the next leap in computer-use agent capabilities. This approach represents a crucial step toward developing truly general computer-use agents, systems that can navigate the digital world with the flexibility and comprehension previously exclusive to human users.

\section{Architecture and System Design}
Following the architecture of AIOS \citep{mei2024aios,rama2025cerebrum}, which abstracts a specific AIOS kernel for isolating agent logic \citep{rama2025cerebrum} and managing agent-related resources and services, we extend this architecture to create \platform, a platform optimized for developing and running computer-use agents.

\begin{figure}[tb!]
    \centering
    \includegraphics[width=1.0\linewidth]{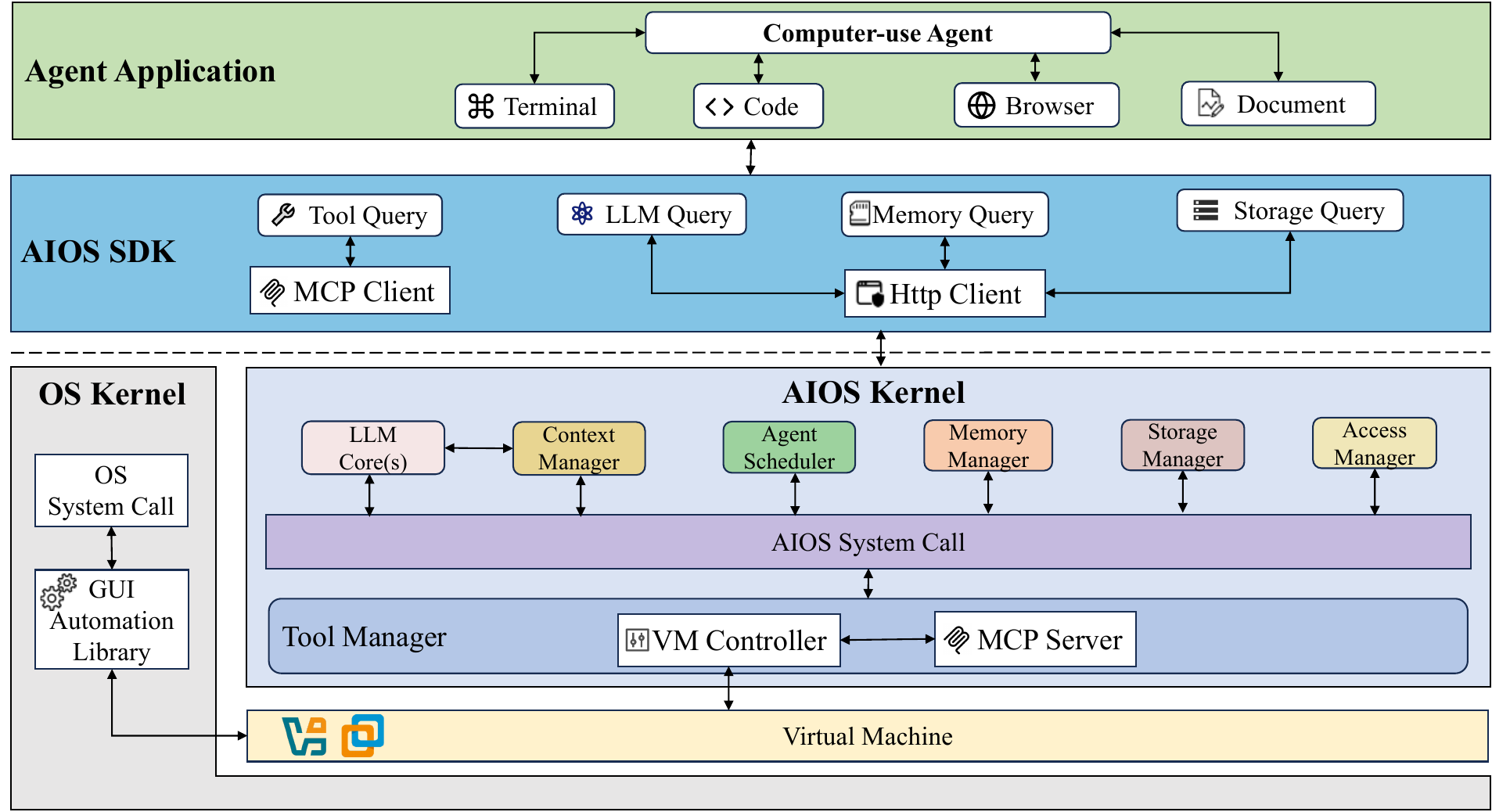}
    \caption{The serving architecture of \platform for computer-use agent, which extends on the basis of AIOS 0.x versions \citep{mei2024aios}. }
    \label{fig:arc}
\end{figure}

\subsection{Architecture}
\textbf{Application Layer:} 
In the Application Layer, agents are provided with SDK APIs to interact with some crucial components like Terminal, Code editor, Browser, and Document. By abstracting the essential interaction patterns common across all interfaces, \platform creates a consistent interaction surface that allows agents to navigate diverse computing environments. This approach significantly reduces the complexity agents face when operating in heterogeneous computer components. 
The agents can leverage the system-level resources, such as LLM and vector database by revoking the APIs within AIOS SDK \citep{rama2025cerebrum}. While maintaining the original design from AIOS SDK regarding LLM, Memory and Storage Query APIs, \platform designs the integration of MCP for tool communications. MCP Client and Http Client are established as communication channels to map the agent's semantic understanding into computer manipulation commands.
    
\textbf{Kernel Layer:} 
\platform extends the AIOS Kernel with significant enhancements focused on computer contextualization. While preserving essential components like LLM Core(s), Context Manager, and Memory Manager, we have fundamentally redesigned the Tool Manager module to incorporate a VM (Virtual Machine) Controller and MCP Server. This redesign creates a sandboxed environment that allows agents to safely interact with computer systems while maintaining a consistent semantic mapping between agent intentions and computer operations. The AIOS system calls are also complimented with Tool related system calls. 

\subsection{Contextualizing Computers as MCP Servers}
The core innovation of \platform lies in its approach to transforming computers from operational tools into contextual environments that language models can naturally comprehend. As shown in \autoref{fig:mcp}, our MCP server serves a semantic bridge between agent cognition and computer operation through a structured contextualization process:

\begin{figure}[tb!]
    \centering
    \includegraphics[width=0.85\linewidth]{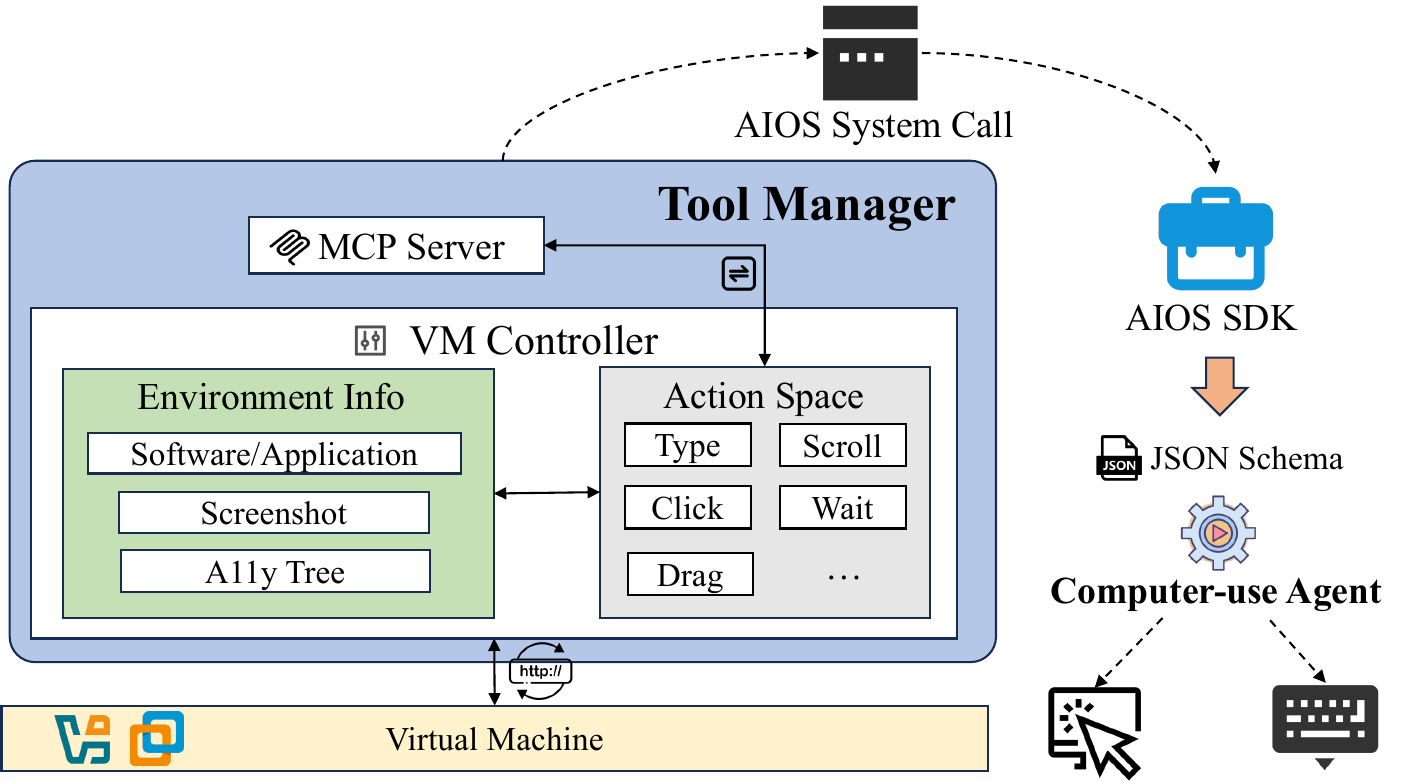}
    \caption{Pipeline for agents to take actions to interact with VM environments. }
    \label{fig:mcp}
\end{figure}

\textbf{Environment Perception Framework:} 
Our environment perception system implements a multi-modal sensing approach that captures the computer state through complementary channels: Screenshot to provide visual signals and Accessibility Tree to provide structural information within the the current screen. Aside from the information from the visible elements in the screenshot, we also provide the mechanisms to inspect invisible information like the version information of softwares. 
Through these components, we provide a comprehensive semantic representation of the computing environment. The perception components allows agents to understand the computer environment as a meaningful semantic landscape rather than a collection of pixels or interface elements, benefiting their ability to reason about computer states and potential actions.
    
\textbf{Action Space Semantics:} 
Within \platform, the action space is constrained into atomic but essential computer-related operations including CLICK, SCROLL, TYPE, DRAG, etc. Each action type is further encoded into the combination of several GUI control signals by leveraging the pyautogui\footnote{https://pyautogui.readthedocs.io/en/latest/} library. This action space enables agents to plan and execute complex interaction sequences while reasoning about operations at a conceptual level similar to human thought processes.

\textbf{VM Controller: }
Considering that some of the computer-use operations can lead to irreversible outcomes, we design the VM controller to allow agents to reason and act in a safe sandbox. The HTTP interface to the Virtual Machine establishes a standardized communication protocol and serves as the abstraction of low-level operations to allow agents to interact with computer environments regardless of the underlying implementation details. 

Through above contextualization, \platform transforms computers from operational challenges into semantic environments that language models can naturally navigate, substantially reducing the cognitive gap between language understanding and computer interaction.

\section{\agent: a Computer-Use Agent built on \platform}

To demonstrate the effectiveness of the \platform platform, we developed \agent, a lightweight computer-use agent that leverages our contextualized environment architecture. We will introduce its design in the following sections. 

\subsection{Orchestrator-Worker Architecture}
\begin{figure}[tb!]
    \centering
    \includegraphics[width=1.0\linewidth]{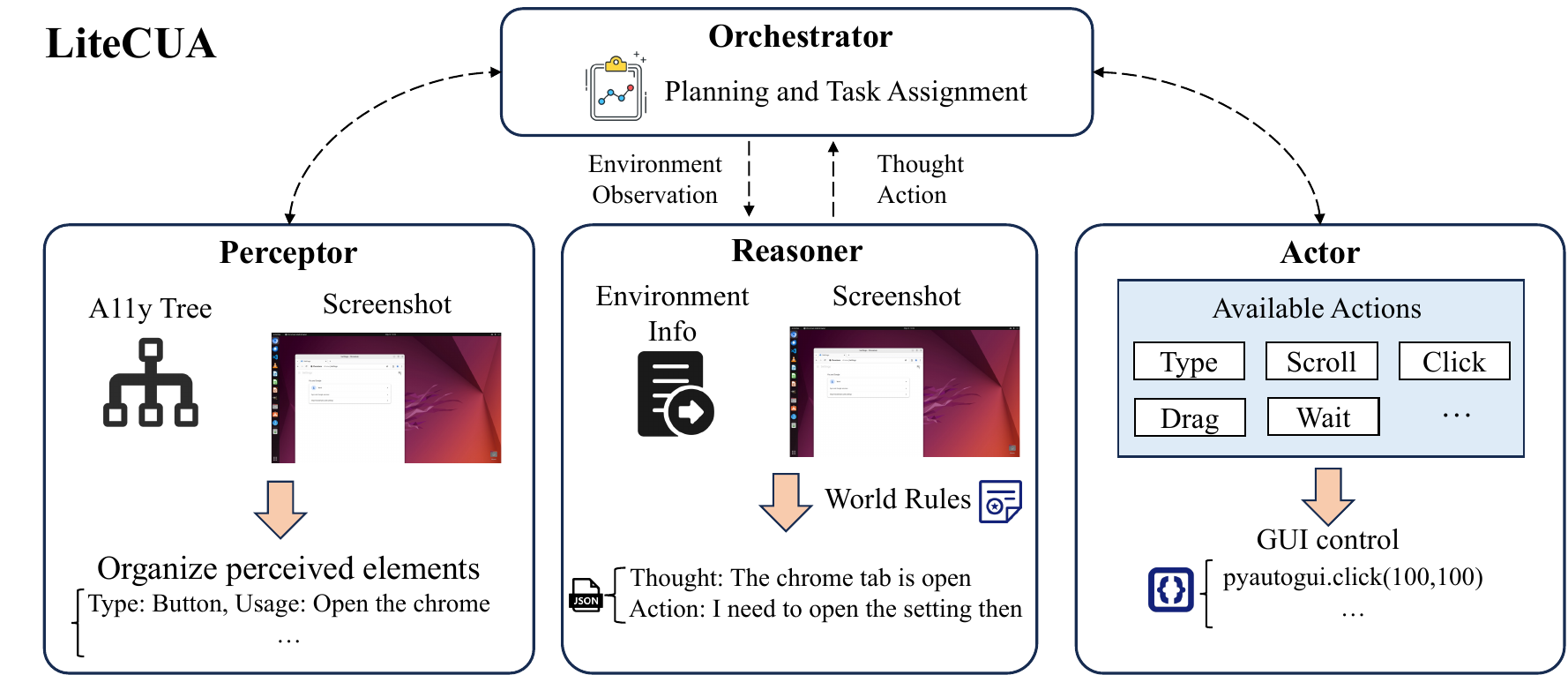}
    \caption{Design of \agent. We adopt a simple orchestrator-worker architecture, where orchestrator is responsible for planning and assigning tasks and other workers deal with specialized duties like perceiving, reasoning and acting. }
    \label{fig:cua}
\end{figure}

As illustrated in \autoref{fig:cua}, \agent adopts an orchestrator-worker architecture that distributes cognitive responsibilities across specialized modules while maintaining cohesive task execution, which is inspired by some popular agent frameworks \citep{wu2023autogen, agashe2025agent, li2023camel}. 

The Orchestrator serves as the central planning and coordination module, responsible for decomposing complex tasks into manageable subtasks, tracking progress toward goal completion. Worker modules perform specialized functions within the agent's cognitive pipeline. This specialization allows each module to optimize for its specific function without conflating distinct cognitive processes. The modular architecture also facilitates incremental improvements to specific capabilities without requiring comprehensive system redesign. 

\subsection{Perceive-Reason-then-Act}
In \agent, we deliberately design a perceive-reason-act cognitive cycle that systematically processes computer environments through three specialized stages:

The \textbf{Perceptor} module integrates multi-modal inputs, i.e., screenshots and A11y tree to construct a comprehensive representation of the computing environment. Rather than processing raw inputs directly, the Perceptor organizes perceived elements into semantically meaningful structures that capture both the functional and relational aspects of interface elements. For example, a button is represented not merely as a visual element but as an interactive component with specific usage semantics, e.g., "Type: Button, Usage: Open the chrome".

The \textbf{Reasoner} module serves as the cognitive core of \agent, interpreting the structured environment information in relation to the current task objectives. It processes textual environment description alongside visual screenshot data, applying operational rules that encode interaction patterns and constraints within computing environments. These information will be fed into the Reasoner to generate both thought (interpretative understanding of the current state) and action (strategic planning for next steps) outputs. 

The \textbf{Actor} module translates high-level action intentions into precise computer operations through a structured action space. It transforms semantic actions (Type, Scroll, Click, Drag, Wait) to specific GUI control commands leveraging pyautogui, e.g., \texttt{pyautogui.click(100,100)}. 

This perceive-reason-act cycle enables \agent to maintain goal-directed behavior across extended interaction sequences while adapting to dynamic computing environments. By leveraging \platform's contextualized environment architecture,  we demonstrate that effective computer-use agents can be constructed with relatively simple and straightforward architectures and the performance of \agent will be shown in Section \ref{sec:eval}. 

\section{Evaluation} \label{sec:eval}

\subsection{Setup} \label{sec:setup}
\paragraph{Deployment Environment.}
The experiments are conducted on an isolated VM with 2 AMD Ryzen CPU cores
and 4 GB of memory with ubuntu 22.04 desktop. 
\paragraph{Benchmark.} We evaluate the \agent computer-use agent built atop of \platform on OSWorld \citep{xie2024osworld}, a benchmark which provides a diverse online environment for evaluating multi-modal agents on computer-using tasks. It consists of 369 tasks involving real-world web and desktop applications. 
\paragraph{Baselines.} For baselines, we choose three LLM baselines, i.e., GPT (4o-mini, GPT-4o) \citep{OpenAIGPT4}, and Gemini-1.5-flash \citep{team2023gemini}, and three agent baselines which adopt GPT-4o as their backbones, i.e., Friday \citep{wu2024copilot}, Open-Interpreter \citep{interpreter2024}, and AgentStore \citep{jia2024agentstore}. For a fair comparison, we evaluate all the methods under the "Screenshot + A11y tree" setting and only compare agents which do not introduce extra models for grounding. The metric we use is success rate provided within OSWorld. 

\subsection{Comparative Performance}

\begin{table}[tb!]
\centering
\begin{minipage}{0.45\linewidth}
\centering
\caption{Success Rate Comparison} \label{tab:overall}
\begin{tabular}{lc}
\toprule
Method (Screenshot + A11y tree)                           & OSWorld \\ \midrule
\multicolumn{2}{c}{LLMs}                  \\ \midrule
GPT-4o-mini \citep{OpenAIGPT4}                     & 6.21    \\
GPT-4o \citep{OpenAIGPT4}                          & 11.21   \\
Gemini-1.5-pro \citep{team2023gemini}                  & 5.1     \\ \midrule
\multicolumn{2}{c}{Agents (based on GPT-4o)}         \\ \midrule
Friday \citep{wu2024copilot}                          & 11.11   \\
Open-Interpreter \citep{interpreter2024}                & 8.94    \\
AgentStore \citep{jia2024agentstore} & 13.55   \\
\textbf{\agent}                       &  \textbf{14.66}       \\
\bottomrule
\end{tabular}
\end{minipage}
\hfill
\begin{minipage}{0.48\linewidth}
\centering
\caption{Detailed Performance of \agent} \label{tab:breakdown}
\begin{tabular}{lcc}
\toprule
Domain                & Score & Average Steps \\
\midrule
Chrome          & 7/46  & 28.4               \\
GIMP            & 4/26  & 32.7               \\
Libreoffice Calc & 0/47  & 50.0               \\
Libreoffice Writer & 2/23 & 42.1               \\
Libreoffice Impress & 7/47 & 31.5               \\
OS              & 13/24 & 18.2               \\
VLC             & 3.11/17 & 35.8               \\
Thunderbird     & 0/15  & 50.0               \\
VSCode          & 8/23  & 25.6               \\
Multi-app       & 11/101 & 38.4               \\
Total           & 54.10/369 & 35.3               \\
\bottomrule
\end{tabular}
\end{minipage}
\end{table}

From \autoref{tab:overall}, \agent achieves a 14.66\% success rate on the OSWorld benchmark, showing modest improvements over both standalone model baselines and other agent frameworks. Compared to GPT-4o (11.21\%), our approach demonstrates that properly contextualizing computer environments for LLMs can yield better results than using the model in isolation. Among agent frameworks using GPT-4o as their backbone, \agent outperforms Friday (11.11\%) and Open-Interpreter (8.94\%), with a slight edge over AgentStore (13.55\%). These results, while still showing significant room for improvement, suggest that our approach of treating computers as contextual environments provides advantages for computer-use tasks. The AIOS Agent Operating System architecture allows for more effective translation between LLM understanding and computer operations compared to traditional prompt-based approaches. However, the overall modest success rates across all systems (below 15\%) indicate that computer-use tasks remain challenging for current AI systems, suggesting that significant architectural and modeling improvements are still needed to achieve robust performance.

\subsection{Performance Breakdown by Domains}
The domain-specific results reveal clear patterns in \agent's capabilities and limitations. The system performs best on OS-related tasks (13/24, 54.2\%) with the lowest average step count (18.2), likely because these operations involve clearer, more structured interactions. Similarly, VSCode tasks show reasonable performance (8/23, 34.8\%) with relatively efficient execution (25.6 steps on average). In contrast, the system struggles significantly with complex productivity applications, achieving no success on Libreoffice Calc (0/47) or Thunderbird (0/15). These domains typically involve more complex screenshots that need to be perceived and have specialized interaction patterns. The maximum step count (50.0) in these domains indicates that the system exhausts its operation budget without achieving the task goals. Multi-application tasks, which require maintaining context across different environments, show modest performance (11/101, 10.9\%) with higher average step counts (38.4), reflecting the additional complexity of context switching. Web and media applications (Chrome and VLC) show intermediate performance. 
% These patterns suggest that \platform's effectiveness correlates with the clarity of the semantic mapping between visual elements and their functions in different applications. 

% \subsection{Error Analysis}

\section{Related Work}

\subsection{Computer-Use Agent}
The automation of computer-based tasks through agents has significantly advanced with (M)LLMs, enabling them to operate across desktop, mobile, and web platforms. A key technical improvement involves moving beyond reliance on potentially noisy textual representations like HTML or accessibility trees towards more robust, vision-centric perception. For instance, systems like UGround \citep{gou2025navigating} and Aria-UI \citep{yang2024ariaui} champion agents that perceive GUIs visually, akin to humans, and perform pixel-level operations. UGround leverages large-scale synthetic web data and adapts the LLaVA architecture for universal visual grounding , while Aria-UI employs a pure-vision approach with a scalable data pipeline to synthesize diverse instruction samples for grounding and incorporates action histories for context-aware reasoning. On mobile platforms, AutoDroid demonstrates a technical advancement by combining LLM commonsense with app-specific knowledge, acquired through automated dynamic analysis and exploration-based memory injection, to automate tasks without prior manual effort \citep{wen2023autodroid}. The evaluation of such computer-use agents is also evolving, with benchmarks like AgentBench \citep{liu2023agentbench}, TheAgentCompany \citep{xu2024agentcompany}, and AgentGym \citep{xi2023agentgym} providing more complex and realistic scenarios for assessing their operational capabilities. Furthermore, the convergence of API-based and GUI-based interaction, creating hybrid agents, is a promising direction for versatile software automation \citep{qiao2025apivsgui}. 

\subsection{AI Agent Operating System}
The increasing complexity of LLM-based agents calls for dedicated infrastructure, leading to the development of AI Agent Operating Systems designed to provide foundational services. AIOS \citep{mei2024aios}, as the first LLM agent operating system, introduces an LLM-specific kernel to manage resources and LLM-specific services, offering agent scheduling, context management, memory and storage management, tool management, and access control, along with the agent SDK \citep{rama2025cerebrum} for agent development. Such systems address critical challenges like efficient resource allocation, a problem also explored by research into self-resource allocation in multi-agent LLM systems where LLMs themselves can act as planners for task distribution \citep{amayuelas2025selfresource}. UFO2 \citep{zhang2025ufo2} designs and implements the desktop agent OS within the Windows environment. These developments show that agent-specific environments are crucial for creating scalable, reliable, and secure ecosystems for advanced agent applications. 

\section{Conclusion and Envision}
In this paper, we introduced \platform, a platform that reconceptualizes the computer-use agent challenge by focusing on environmental contextualization rather than model capabilities alone by contextualizing the computer as the MCP server. We also introduce \agent, our demonstrative agent built on the \platform platform, which achieved competitive performance on the OSWorld benchmark with a 14.66\% success rate, outperforming several specialized agent frameworks despite its relatively simple architecture. 

While our current implementation shows promising results, several directions for future work remain. First, enhancing the perception framework to better capture temporal and causal relationships within computer environments could improve agent performance on complex multi-step tasks. Second, developing more sophisticated action space semantics that incorporate probabilistic reasoning could enable more robust operation in uncertain environments. Finally, extending our approach to additional computing domains, particularly those with specialized interfaces like professional creative applications, could further validate the generalizability of our contextualization methodology. \platform represents a step toward artificial general intelligence systems that can interact with the digital world, not by painstakingly teaching AI to use interfaces designed for humans, but by transforming those interfaces into environments that AI can naturally comprehend. This paradigm shift, from adapting agents to computers toward adapting computers for agents, offers a promising path forward for developing truly general computer-use capabilities in AI systems.

\bibliographystyle{plain}
\bibliography{aios}

\begin{thebibliography}{10}

\bibitem{OpenAIGPT4}
Josh Achiam, Steven Adler, Sandhini Agarwal, Lama Ahmad, Ilge Akkaya, Florencia~Leoni Aleman, Diogo Almeida, Janko Altenschmidt, Sam Altman, Shyamal Anadkat, et~al.
\newblock Gpt-4 technical report.
\newblock {\em arXiv preprint arXiv:2303.08774}, 2023.

\bibitem{agashe2025agent}
Saaket Agashe, Kyle Wong, Vincent Tu, Jiachen Yang, Ang Li, and Xin~Eric Wang.
\newblock Agent s2: A compositional generalist-specialist framework for computer use agents.
\newblock {\em arXiv preprint arXiv:2504.00906}, 2025.

\bibitem{amayuelas2025selfresource}
Alfonso Amayuelas, Jingbo Yang, Saaket Agashe, Ashwin Nagarajan, Antonis Antoniades, Xin~Eric Wang, and William Wang.
\newblock Self-resource allocation in multi-agent {LLM} systems.
\newblock {\em arXiv preprint arXiv:2504.02051}, Apr 2025.

\bibitem{besta2023graph}
Maciej Besta, Nils Blach, Ales Kubicek, Robert Gerstenberger, Lukas Gianinazzi, Joanna Gajda, Tomasz Lehmann, Michal Podstawski, Hubert Niewiadomski, Piotr Nyczyk, et~al.
\newblock Graph of thoughts: Solving elaborate problems with large language models.
\newblock {\em arXiv preprint arXiv:2308.09687}, 2023.

\bibitem{bonatti2024windows}
Rogerio Bonatti, Dan Zhao, Francesco Bonacci, Dillon Dupont, Sara Abdali, Yinheng Li, Yadong Lu, Justin Wagle, Kazuhito Koishida, Arthur Bucker, et~al.
\newblock Windows agent arena: Evaluating multi-modal os agents at scale.
\newblock {\em arXiv preprint arXiv:2409.08264}, 2024.

\bibitem{chen2022program}
Wenhu Chen, Xueguang Ma, Xinyi Wang, and William~W Cohen.
\newblock Program of thoughts prompting: Disentangling computation from reasoning for numerical reasoning tasks.
\newblock {\em arXiv preprint arXiv:2211.12588}, 2022.

\bibitem{deng2024mobile}
Shihan Deng, Weikai Xu, Hongda Sun, Wei Liu, Tao Tan, Jianfeng Liu, Ang Li, Jian Luan, Bin Wang, Rui Yan, et~al.
\newblock Mobile-bench: An evaluation benchmark for llm-based mobile agents.
\newblock {\em arXiv preprint arXiv:2407.00993}, 2024.

\bibitem{deng2023mind2web}
Xiang Deng, Yu~Gu, Boyuan Zheng, Shijie Chen, Samuel Stevens, Boshi Wang, Huan Sun, and Yu~Su.
\newblock Mind2web: Towards a generalist agent for the web.
\newblock {\em Advances in Neural Information Processing Systems}, 36, 2023.

\bibitem{dubey2024llama}
Abhimanyu Dubey, Abhinav Jauhri, Abhinav Pandey, Abhishek Kadian, Ahmad Al-Dahle, Aiesha Letman, Akhil Mathur, Alan Schelten, Amy Yang, Angela Fan, et~al.
\newblock The llama 3 herd of models.
\newblock {\em arXiv preprint arXiv:2407.21783}, 2024.

\bibitem{ge2023openagi}
Yingqiang Ge, Wenyue Hua, Kai Mei, Juntao Tan, Shuyuan Xu, Zelong Li, and Yongfeng Zhang.
\newblock {OpenAGI: When LLM Meets Domain Experts}.
\newblock {\em Advances in Neural Information Processing Systems}, 36, 2023.

\bibitem{gou2025navigating}
Boyu Gou, Ruohan Wang, Boyuan Zheng, Yanan Xie, Cheng Chang, Yiheng Shu, Huan Sun, and Yu~Su.
\newblock Navigating the digital world as humans do: Universal visual grounding for {GUI} agents.
\newblock In {\em International Conference on Learning Representations (ICLR)}, 2025.
\newblock arXiv preprint arXiv:2410.05243 (2024).

\bibitem{guo2025seed1}
Dong Guo, Faming Wu, Feida Zhu, Fuxing Leng, Guang Shi, Haobin Chen, Haoqi Fan, Jian Wang, Jianyu Jiang, Jiawei Wang, et~al.
\newblock Seed1. 5-vl technical report.
\newblock {\em arXiv preprint arXiv:2505.07062}, 2025.

\bibitem{jia2024agentstore}
Chengyou Jia, Minnan Luo, Zhuohang Dang, Qiushi Sun, Fangzhi Xu, Junlin Hu, Tianbao Xie, and Zhiyong Wu.
\newblock Agentstore: Scalable integration of heterogeneous agents as specialized generalist computer assistant.
\newblock {\em arXiv preprint arXiv:2410.18603}, 2024.

\bibitem{jin2024learning}
Can Jin, Tong Che, Hongwu Peng, Yiyuan Li, Dimitris~N. Metaxas, and Marco Pavone.
\newblock Learning from teaching regularization: Generalizable correlations should be easy to imitate.
\newblock In {\em The Thirty-eighth Annual Conference on Neural Information Processing Systems}, 2024.

\bibitem{jin2025two}
Can Jin, Hongwu Peng, Qixin Zhang, Yujin Tang, Dimitris~N Metaxas, and Tong Che.
\newblock Two heads are better than one: Test-time scaling of multi-agent collaborative reasoning.
\newblock {\em arXiv preprint arXiv:2504.09772}, 2025.

\bibitem{jin2024impact}
Mingyu Jin, Qinkai Yu, Dong Shu, Haiyan Zhao, Wenyue Hua, Yanda Meng, Yongfeng Zhang, and Mengnan Du.
\newblock The impact of reasoning step length on large language models.
\newblock {\em arXiv preprint arXiv:2401.04925}, 2024.

\bibitem{koh2024visualwebarena}
Jing~Yu Koh, Robert Lo, Lawrence Jang, Vikram Duvvur, Ming~Chong Lim, Po-Yu Huang, Graham Neubig, Shuyan Zhou, Ruslan Salakhutdinov, and Daniel Fried.
\newblock Visualwebarena: Evaluating multimodal agents on realistic visual web tasks.
\newblock {\em arXiv preprint arXiv:2401.13649}, 2024.

\bibitem{li2023camel}
Guohao Li, Hasan Hammoud, Hani Itani, Dmitrii Khizbullin, and Bernard Ghanem.
\newblock Camel: Communicative agents for "mind" exploration of large language model society.
\newblock {\em Advances in Neural Information Processing Systems}, 36, 2023.

\bibitem{liu2024deepseek}
Aixin Liu, Bei Feng, Bing Xue, Bingxuan Wang, Bochao Wu, Chengda Lu, Chenggang Zhao, Chengqi Deng, Chenyu Zhang, Chong Ruan, et~al.
\newblock Deepseek-v3 technical report.
\newblock {\em arXiv preprint arXiv:2412.19437}, 2024.

\bibitem{liu2023agentbench}
Xiao Liu, Hao Yu, Hanchen Zhang, Yaran Xu, Xuanyu Du, Xian Li, Hongyu Wang, Yaqing Chang, Yifei Zhao, Yao Liu, et~al.
\newblock {AgentBench: Evaluating LLMs as Agents}.
\newblock {\em arXiv preprint arXiv:2308.03688}, Aug 2023.

\bibitem{interpreter2024}
Killian Lucas.
\newblock Open interpreter.
\newblock \url{https://github.com/OpenInterpreter/open-interpreter}, 2024.

\bibitem{mei2024aios}
Kai Mei, Xi~Zhu, Wujiang Xu, Wenyue Hua, Mingyu Jin, Zelong Li, Shuyuan Xu, Ruosong Ye, Yingqiang Ge, and Yongfeng Zhang.
\newblock {AIOS: LLM Agent Operating System}.
\newblock {\em arXiv preprint arXiv:2403.16971}, Mar 2024.

\bibitem{pan2024webcanvas}
Yichen Pan, Dehan Kong, Sida Zhou, Cheng Cui, Yifei Leng, Bing Jiang, Hangyu Liu, Yanyi Shang, Shuyan Zhou, Tongshuang Wu, et~al.
\newblock Webcanvas: Benchmarking web agents in online environments.
\newblock {\em arXiv preprint arXiv:2406.12373}, 2024.

\bibitem{patil2024gorilla}
Shishir~G Patil, Tianjun Zhang, Xin Wang, and Joseph~E Gonzalez.
\newblock Gorilla: Large language model connected with massive apis.
\newblock {\em Advances in Neural Information Processing Systems}, 37:126544--126565, 2024.

\bibitem{qiao2025apivsgui}
Yihong Qiao, Zelin Li, Yifei Wang, Shiyao Wang, Ming Scala, Wen-Tau Yih, and Lichan Hong.
\newblock {API-Based or GUI-Based? A Comparative Study of LLM Agents for Software Automation}.
\newblock {\em arXiv preprint arXiv:2503.11069}, Mar 2025.

\bibitem{qin2023toolllm}
Yujia Qin, Shihao Liang, Yining Ye, Kunlun Zhu, Lan Yan, Yaxi Lu, Yankai Lin, Xin Cong, Xiangru Tang, Bill Qian, et~al.
\newblock Toolllm: Facilitating large language models to master 16000+ real-world apis.
\newblock {\em ICLR}, 2024.

\bibitem{qin2025ui}
Yujia Qin, Yining Ye, Junjie Fang, Haoming Wang, Shihao Liang, Shizuo Tian, Junda Zhang, Jiahao Li, Yunxin Li, Shijue Huang, et~al.
\newblock Ui-tars: Pioneering automated gui interaction with native agents.
\newblock {\em arXiv preprint arXiv:2501.12326}, 2025.

\bibitem{rama2025cerebrum}
Balaji Rama, Kai Mei, and Yongfeng Zhang.
\newblock {Cerebrum (AIOS SDK): A Platform for Agent Development, Deployment, Distribution, and Discovery}.
\newblock In {\em Proceedings of the 2025 Conference of the Nations of the Americas Chapter of the Association for Computational Linguistics: Human Language Technologies (System Demonstrations)}, pages 135--142, 2025.

\bibitem{rawles2024androidworld}
Christopher Rawles, Sarah Clinckemaillie, Yifan Chang, Jonathan Waltz, Gabrielle Lau, Marybeth Fair, Alice Li, William Bishop, Wei Li, Folawiyo Campbell-Ajala, et~al.
\newblock Androidworld: A dynamic benchmarking environment for autonomous agents.
\newblock {\em arXiv preprint arXiv:2405.14573}, 2024.

\bibitem{shen2023hugginggpt}
Yongliang Shen, Kaitao Song, Xu~Tan, Dongsheng Li, Weiming Lu, and Yueting Zhuang.
\newblock Hugginggpt: Solving ai tasks with chatgpt and its friends in hugging face.
\newblock {\em Advances in Neural Information Processing Systems}, 36:38154--38180, 2023.

\bibitem{team2023gemini}
Gemini Team, Rohan Anil, Sebastian Borgeaud, Yonghui Wu, Jean-Baptiste Alayrac, Jiahui Yu, Radu Soricut, Johan Schalkwyk, Andrew~M Dai, Anja Hauth, et~al.
\newblock Gemini: a family of highly capable multimodal models.
\newblock {\em arXiv preprint arXiv:2312.11805}, 2023.

\bibitem{wang2024oscar}
Xiaoqiang Wang and Bang Liu.
\newblock Oscar: Operating system control via state-aware reasoning and re-planning.
\newblock {\em arXiv preprint arXiv:2410.18963}, 2024.

\bibitem{wang2024executable}
Xingyao Wang, Yangyi Chen, Lifan Yuan, Yizhe Zhang, Yunzhu Li, Hao Peng, and Heng Ji.
\newblock Executable code actions elicit better llm agents.
\newblock In {\em Forty-first International Conference on Machine Learning}, 2024.

\bibitem{wei2023chainofthought}
Jason Wei, Xuezhi Wang, Dale Schuurmans, Maarten Bosma, Fei Xia, Ed~Chi, Quoc~V Le, Denny Zhou, et~al.
\newblock Chain-of-thought prompting elicits reasoning in large language models.
\newblock {\em Advances in Neural Information Processing Systems}, 35:24824--24837, 2022.

\bibitem{wen2023autodroid}
Hao Wen, Yuanchun Li, Guohong Liu, Shanhui Zhao, Tao Yu, Toby Jia-Jun Li, Shiqi Jiang, Yunhao Liu, Yaqin Zhang, and Yunxin Liu.
\newblock {AutoDroid: LLM-powered} task automation in android.
\newblock In {\em Proceedings of the Annual International Conference on Mobile Computing and Networking (MobiCom)}, 2024.
\newblock arXiv preprint arXiv:2308.15272 (2023).

\bibitem{wu2023autogen}
Qingyun Wu, Gagan Bansal, Jieyu Zhang, Yiran Wu, Shaokun Zhang, Erkang Zhu, Beibin Li, Li~Jiang, Xiaoyun Zhang, and Chi Wang.
\newblock Autogen: Enabling next-gen llm applications via multi-agent conversation framework.
\newblock {\em arXiv preprint arXiv:2308.08155}, 2023.

\bibitem{wu2024copilot}
Zhiyong Wu, Chengcheng Han, Zichen Ding, Zhenmin Weng, Zhoumianze Liu, Shunyu Yao, Tao Yu, and Lingpeng Kong.
\newblock Os-copilot: Towards generalist computer agents with self-improvement.
\newblock {\em arXiv preprint arXiv:2402.07456}, 2024.

\bibitem{xi2023agentgym}
Zhisheng Xi, Yaofei Chen, Yuan Liu, Tianyu Zhang, Fei Wang, Weidong Chen, and Kai Huang.
\newblock {AgentGym: Evolving Large Language Model-based Agents across Diverse Environments}.
\newblock {\em arXiv preprint arXiv:2312.08182}, Dec 2023.

\bibitem{xie2024osworld}
Tianbao Xie, Danyang Zhang, Jixuan Chen, Xiaochuan Li, Siheng Zhao, Ruisheng Cao, Toh~J Hua, Zhoujun Cheng, Dongchan Shin, Fangyu Lei, et~al.
\newblock Osworld: Benchmarking multimodal agents for open-ended tasks in real computer environments.
\newblock {\em Advances in Neural Information Processing Systems}, 37:52040--52094, 2024.

\bibitem{xu2024agentcompany}
Michihiro Xu, Jewell Zhao, Justin GerLach, and Xin Wang.
\newblock {TheAgentCompany: A Gamified Benchmark to Evaluate AI Agents in a Simulated Software Company}.
\newblock {\em arXiv preprint arXiv:2402.00811}, Feb 2024.

\bibitem{xu2023tool}
Qiantong Xu, Fenglu Hong, Bo~Li, Changran Hu, Zhengyu Chen, and Jian Zhang.
\newblock On the tool manipulation capability of open-source large language models.
\newblock {\em arXiv preprint arXiv:2305.16504}, 2023.

\bibitem{yang2024qwen2}
An~Yang, Baosong Yang, Beichen Zhang, Binyuan Hui, Bo~Zheng, Bowen Yu, Chengyuan Li, Dayiheng Liu, Fei Huang, Haoran Wei, et~al.
\newblock Qwen2. 5 technical report.
\newblock {\em arXiv preprint arXiv:2412.15115}, 2024.

\bibitem{yangswe}
John Yang, Carlos~E Jimenez, Alexander Wettig, Kilian Lieret, Shunyu Yao, Karthik Narasimhan, and Ofir Press.
\newblock Swe-agent: Agent-computer interfaces enable automated software engineering.

\bibitem{yang2024ariaui}
Yuhao Yang, Yue Wang, Dongxu Li, Ziyang Luo, Bei Chen, Chao Huang, and Junnan Li.
\newblock {Aria-UI: Visual Grounding for GUI Instructions}.
\newblock {\em arXiv preprint arXiv:2412.16256}, Dec 2024.

\bibitem{yao2022webshop}
Shunyu Yao, Howard Chen, John Yang, and Karthik Narasimhan.
\newblock Webshop: Towards scalable real-world web interaction with grounded language agents.
\newblock {\em Advances in Neural Information Processing Systems}, 35:20744--20757, 2022.

\bibitem{yao2023tree}
Shunyu Yao, Dian Yu, Jeffrey Zhao, Izhak Shafran, Thomas~L Griffiths, Yuan Cao, and Karthik Narasimhan.
\newblock Tree of thoughts: Deliberate problem solving with large language models.
\newblock {\em arXiv preprint arXiv:2305.10601}, 2023.

\bibitem{yao2023react}
Shunyu Yao, Jeffrey Zhao, Dian Yu, Nan Du, Izhak Shafran, Karthik Narasimhan, and Yuan Cao.
\newblock {ReAct}: Synergizing reasoning and acting in language models.
\newblock {\em International Conference on Learning Representations}, 2023.

\bibitem{yu2024exact}
Xiao Yu, Baolin Peng, Vineeth Vajipey, Hao Cheng, Michel Galley, Jianfeng Gao, and Zhou Yu.
\newblock Exact: Teaching ai agents to explore with reflective-mcts and exploratory learning.
\newblock {\em arXiv preprint arXiv:2410.02052}, 2024.

\bibitem{zhang2025ufo2}
Chaoyun Zhang, He~Huang, Chiming Ni, Jian Mu, Si~Qin, Shilin He, Lu~Wang, Fangkai Yang, Pu~Zhao, Chao Du, et~al.
\newblock Ufo2: The desktop agentos.
\newblock {\em arXiv preprint arXiv:2504.14603}, 2025.

\bibitem{zhang2024codeagent}
Kechi Zhang, Jia Li, Ge~Li, Xianjie Shi, and Zhi Jin.
\newblock Codeagent: Enhancing code generation with tool-integrated agent systems for real-world repo-level coding challenges.
\newblock {\em arXiv preprint arXiv:2401.07339}, 2024.

\bibitem{zhang2025enhancing}
Zhisong Zhang, Tianqing Fang, Kaixin Ma, Wenhao Yu, Hongming Zhang, Haitao Mi, and Dong Yu.
\newblock Enhancing web agents with explicit rollback mechanisms.
\newblock {\em arXiv preprint arXiv:2504.11788}, 2025.

\bibitem{zhou2023webarena}
Shuyan Zhou, Frank~F Xu, Hao Zhu, Xuhui Zhou, Robert Lo, Abishek Sridhar, Xianyi Cheng, Tianyue Ou, Yonatan Bisk, Daniel Fried, et~al.
\newblock Webarena: A realistic web environment for building autonomous agents.
\newblock {\em arXiv preprint arXiv:2307.13854}, 2023.

\end{thebibliography}

\end{document}